# Biological Organisms as End Effectors

Josephine Galipon[1,2,3], Shoya Shimizu[3], Kenjiro Tadakuma[3,4,*]

[1]Graduate School of Science and Engineering, Yamagata University, Japan

[2]Institute for Advanced Biosciences, Keio University, Japan

[3]Graduate School of Information Sciences, Tohoku University, Japan

[4]Tough Cyberphysical AI Research Center, Tohoku University, Japan

*corresponding author: tadakuma@rm.is.tohoku.ac.jp

## Graphical Abstract

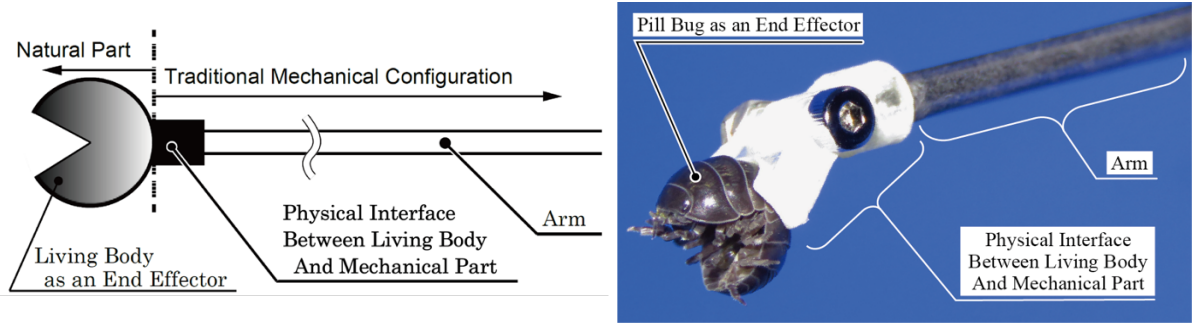

## Abstract

In robotics, an end effector is a device at the end of a robotic arm that is designed to physically interact with objects in the environment or with the environment itself. Effectively, it serves as the hand of the robot, carrying out tasks on behalf of humans. But could we turn this concept on its head and consider using living organisms themselves as end effectors? This paper introduces a novel idea of using whole living organisms as end effectors for robotics. We showcase this by demonstrating that pill bugs and chitons—types of small, harmless creatures—can be utilized as functional grippers. Crucially, this method does not harm these creatures, enabling their release back into nature after use. How this concept may be expanded to other organisms and applications is also discussed.

## Introduction

The field of biorobotics may be arbitrarily divided into classical biorobotics or biomimetics, where the robot is a model for studying a living organism—also known as the *understanding by building* approach—and interactive biorobotics, where the robot interacts with the living organism to gain information about its cognition and behavior [1]. Within the latter categorization, the interface between robotic devices and living organisms has been explored for decades, not only as a means to either study living organisms, but also to augment their abilities as cyborgs.

The field of hybrid robotics may be divided into subcategories defined by the level of complexity of the living parts based on **Figure 1**, based on the standpoint of engineering. Cells are the smallest unit that can make copies of itself on their own. Cultured neurons may be connected to a virtual or real robotic component to form an animat or hybrot [2]. A combination of cells joined together into a cell sheet or cultured tissues may also be used as hybrid robotic parts, such as cultured cardiomyocytes as a bio-microactuator to drive polymer micropillars [3], origami folding structures containing pig intestine walls to patch stomach wounds *in vivo* [4], and engineered skeletal muscle tissue used for object manipulation by a biohybrid robot [5]. The next level in the hierarchy of biological complexity are whole inner organs and limbs, as demonstrated by our own research applying a sheep intestine as an edible balloon actuator [6].

Finally, at the top level of complexity, whole organisms may also be used as biohybrid devices, as shown in **Figure 2**. Examples of previous research in this area may be divided into three main categories depending on purpose: for remote-sensing and locomotion, for scientific investigation, and as end effectors. The first category includes remote-controlled beetle flight [7], and the steering of cockroaches [8], whereas the second category includes fly-ball and ant-ball tracking systems [9, 10], and more recently the driving of a robotic device by a living silkmoth [11, 12], both of which enable quantitative spatial measurements of insect neural activity. Electricity generated by insects may also be extracted as a source of energy, further enhancing the space of possibilities for sustained interactions between robotic parts and living organisms in hybrid systems [13, 14]. As a side note, it should be mentioned that unicellular organisms (bacteria, some algae and fungi, protozoans) or partially unicellular organisms (slime mold) can be included in both the "cells" and "whole organisms" categories as per **Figure 1**. One example of biohybrid device used a slime mold to power a heart rate sensor inside a smart watch [15].

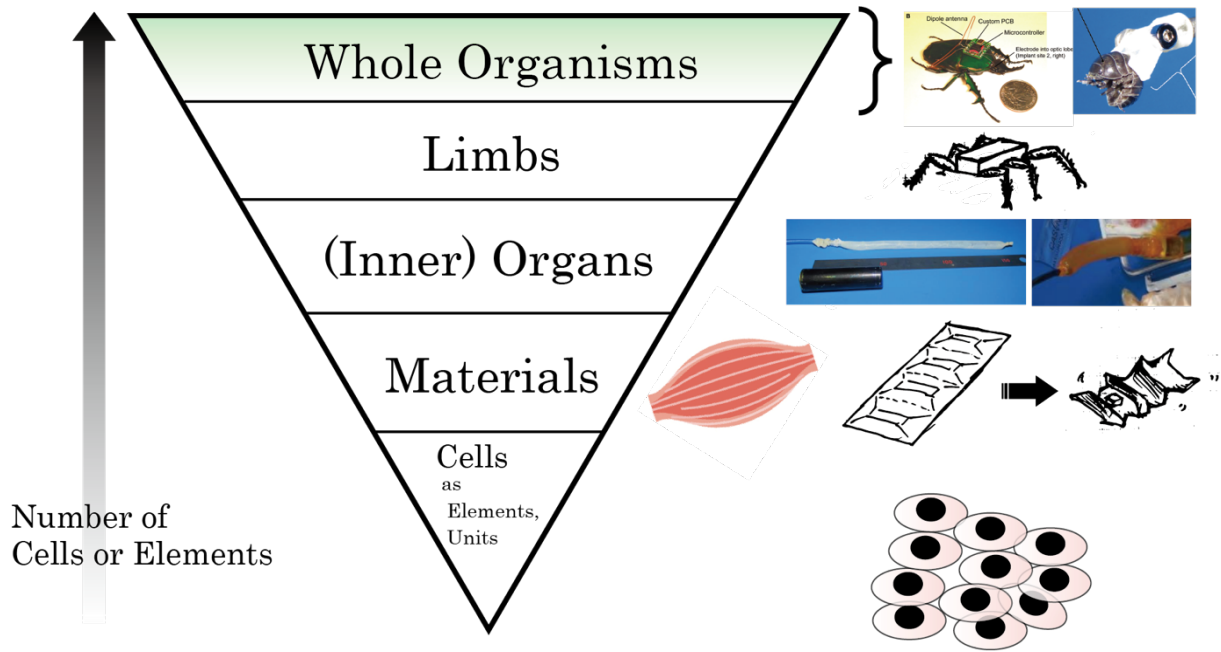

**Figure 1**. Positioning of our research within biohybrid robotics. The beetle for "Whole Organisms" comes from [7]; the drawing for "Limbs" was drawn by us but inspired from the work of Prof. Takeuchi as reviewed in [16]; two images for (Inner) Organs come from [6], the drawing for "Materials" was drawn by us but inspired from [4].

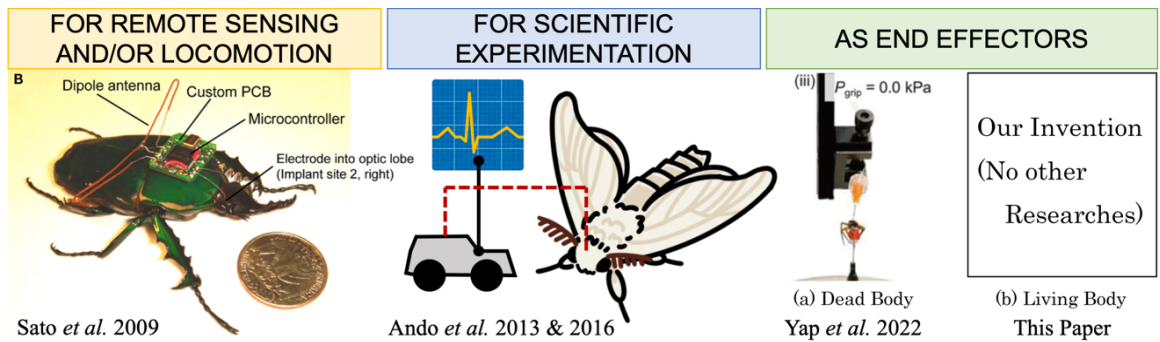

**Figure 2**. Research using whole biological organisms. The image of a moth is from TogoTV (© 2016 DBCLS TogoTV, CC-BY-4.0 https://creativecommons.org/licenses/by/4.0/deed.ja)

Our research further expands on using natural whole bodies in biohybrid robots. Although dead spiders were recently used as gripping actuators in a concept coined as "necrobotics" [17], to our knowledge, there is no prior example of whole living organisms being used as end effectors for robotic arms, which we propose here in **Figure 3**. In this study, we explore the use of the inherent structures and movements of whole living organisms—in particular, those harboring exoskeletons and reflexive movements—as a basis for biohybrid robotic function. This approach departs from traditional methodologies by leveraging the structures and movements of specific body parts without disconnecting them from the organism, all the while preserving the life and integrity of the creature.

As a demonstration of this concept, in this paper we present two end effectors, one exploiting the reflexive closure of a land animal, the pill bug (family Armadillidiidae), and the other taking advantage of the strong suction capability of an aquatic animal, a marine mollusk known as the chiton (class Polyplacophora). Both of these animals' natural behaviors are ideal for grasping objects by exploiting our knowledge of their behavior. As **Figure 3** illustrates, we developed a physical interface that bridges conventional mechanical elements with living organisms, thereby integrating parts of the organism's structure and movement into our robotic design.

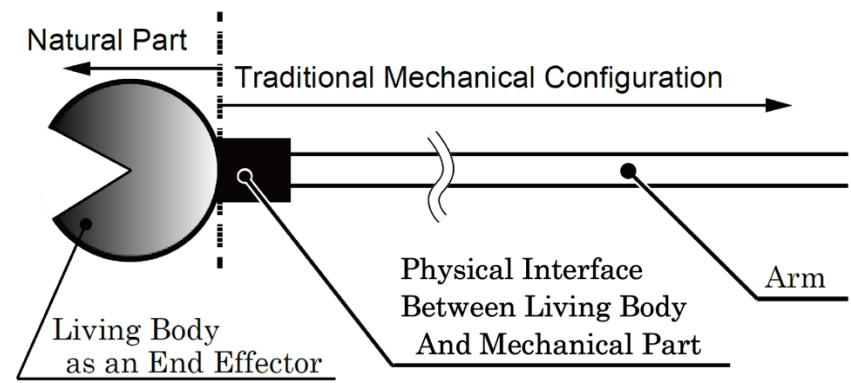

**Figure 3.** Novel concept of living body as end effector. Figure from our conference paper [18].

Considering the range of possible changes in physical parameters and morphology by biological structures and organisms, we believe the manipulation of various objects in a given environment to be feasible in cooperation with living organisms. For instance, we could exploit the locomotion of slugs or the suction capabilities of leeches by affixing them to the edge of our mechanical design. Just as we have been making use of sled dogs and horses for transportation and carrier pigeons to convey postage for thousands of years, the concept proposed here is yet another way of harnessing the unique functions of biological organisms, and is one step further towards a truly radical integration of robots and their environment.

**Methods**

*Construction of the mechanical parts*

The mechanical parts were designed in Pro-Engineering Wildfire 3.0 and 3D printed with AR-M2 ink on an Agilista 3100 printer or with carbon fiber composite on a Markforged MarkTwo printer for the pill bug and chiton modules, respectively. The parts were connected using appropriately sized nuts and bolts.

*Collection and attachment of the biological organisms*

The pill bug used in this study was found on Tohoku University's Aobayama campus, Miyagi prefecture, Sendai City, Japan and released after the experiment. The chiton was collected from the Sea of Japan, Yamagata Prefecture, Shonai area, transported in sea water and kept in 40L seawater with mechanical filtration at ambient temperature for further study.

**Results**

*Building the pill bug gripper prototype model*

In line with the aforementioned concept, we designed and built a functioning device. The pill bug's main body was connected to a 3mm-diameter arm through a physical interface. To allow for the pill bug's reflexive movement post-attachment, we designed a harness-like attachment using a flexible string. With just one fixation line, the pill bug may close up into a ball when it senses a threat (**Figure 4A**), whereas with two fixation lines, the pill bug is stuck the open configuration (**Figure 4B**). **Figure 4C** provides a comprehensive view of the completed device. The width of the biological component was 6.9 mm. When observed laterally in the closed configuration, the cross-sectional diameter of the device is approximately 7.03 mm. The total weight of the end-effector, including the physical interface, is 0.76 g. The parts were custom-made with a 3D printer and connected with nuts and bolts.

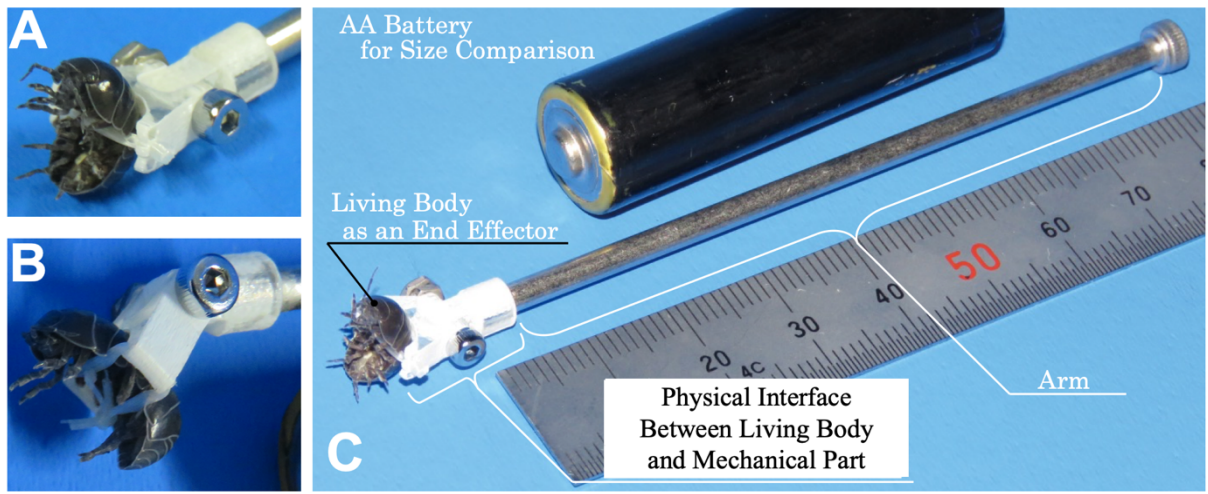

**Figure 4**. Configuration of the pill bug gripper real prototype model. [A] configuration with one fixation line; [B] configuration with two fixation lines; [C] whole view of the entire pill bug gripper. Figures from our conference paper [18].

### *The pill bug gripper is able to grip and release*

**Figure 5** shows the grasping operation of the prototype. The object to be grasped in this experiment is a lightweight piece of cotton (0.03 g). As shown in **Figure 5**, we confirmed that the object can be grasped by the reflexive closing motion of the pill bug. At present, the reflex action of the pill bug closing its shell is stimulated by the contact with the object itself. After about 115 seconds of holding, the object was released naturally. Although the control of the low stimulus and timing to open the shell is a future issue, the possibility of grasping by opening and closing the shell was demonstrated if the object was relatively lightweight and could be held by the animal.

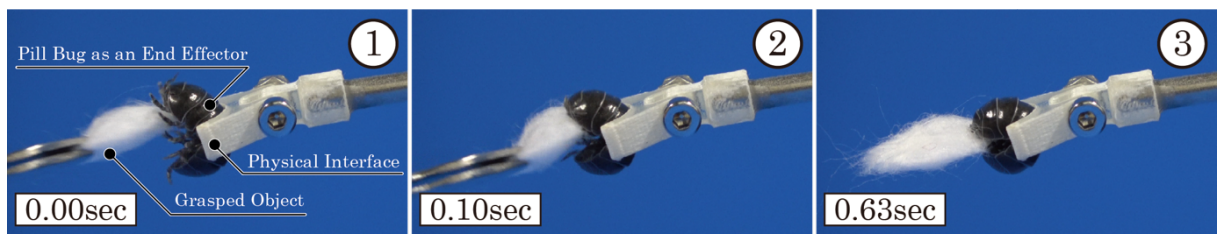

**Figure 5**. Grasping operation using the pill bug gripper. Figure from our conference paper [18].

### *The pill bug end effector is able to manipulate lightweight objects*

As shown in **Figure 6**, the walking motion of the pill bug demonstrated the possibility of realizing an object-handling function. The leg movements are shown in the forward-to-backward direction. Future work will focus on moving lightweight objects in arbitrary directions by turning and controlling them. A video of the pill bug gripper in action is available on Youtube (https://youtu.be/yo_mXCJRFZs) and as a Supplementary file.

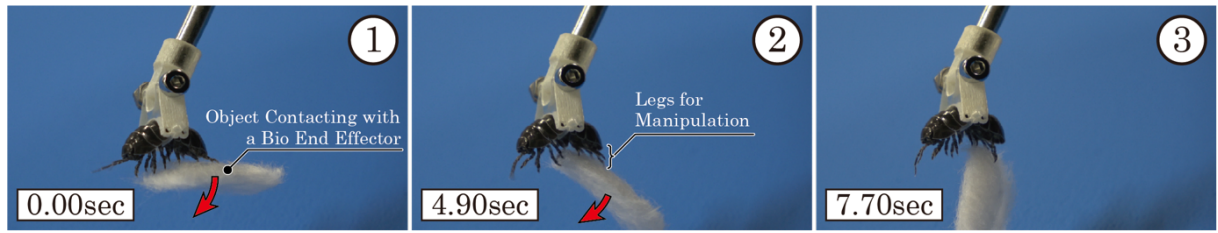

**Figure 6**. Object handling operation using the pill bug gripper. Figure from our conference paper [18].

### Building the chiton gripper prototype model

To accommodate the need for grasping objects underwater, we built a gripper using a marine mollusk, the chiton, known for its ability to strongly attach to rocks and other rough surfaces by a suction mechanism. The chiton used here was 21 mm long at its longest point (18 mm without the girdle) and 12 mm wide at its widest point (9 mm without the girdle) with a height of 6 mm at its highest. A holder was designed to fit the curvature of the chiton's exoskeleton. The holder was fixed to the chiton outside of the water using high-speed epoxy glue. After trial and error we found that it is crucial to thoroughly rub the exoskeleton with 99.5 % alcohol using a Q-tip and let it dry off to remove any substances that may interfere with the glue. We waited 10 minutes for the glue to set. In the meantime, we provided sea water drop by drop to the chiton's front and back sides to prevent the animal from drying up inside, while being careful not to let sea water reach the glue during setting time. The mechanical parts of the robot arm were then attached using a nut and bolt.

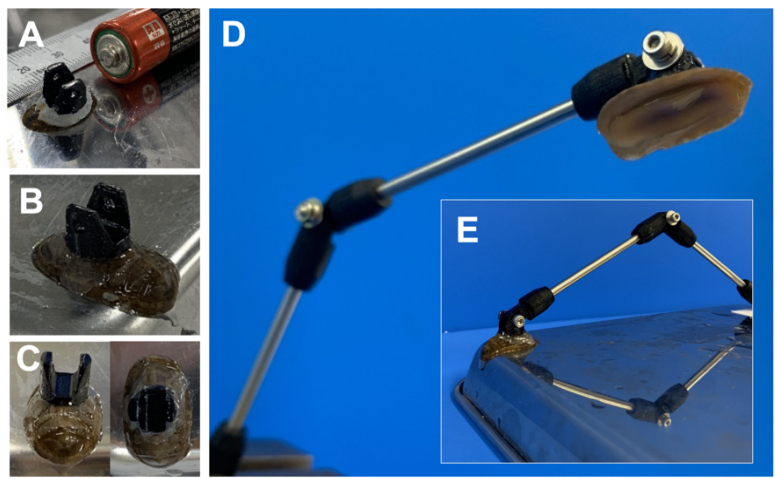

**Figure 7.** Configuration of the chiton gripper real prototype model. [A] a first failed attempt to attach the adapter unit to the chiton using epoxy putty; [B] successful attempt using high-speed epoxy after ethanol

scrubbing; [C] lateral and top view of the chiton with the adapter unit; [D] whole view of the chiton gripper; [E] chiton attaching to a surface showing it is still alive after the operation.

### *Chiton gripping operation successful on various surfaces*

As shown in **Figure 7E** and in **Figure 8** the chiton readily attempts to grip any object that is presented to it, and the object may be lifted provided that the surface properties and weight of the object are within the capabilities of the chiton. We successfully tested the gripping of cork, wooden, and plastic objects. Unlike the chiton, ordinary suction cups cannot attach to cork and wood. Further work on the biology side is needed to establish the upper limits of weight for various surfaces that can be sustained by chitons of various sizes and species.

### *Chiton end effector enables manipulation of objects underwater*

Next, we filmed the chiton underwater as it grasped the object. As the chiton attempted to travel along the object, since the chiton itself is fixed, the object moved accordingly. In **Figure 8**, we can see the clear rotation of a wooden cylinder (25.1 g) and that of a plastic cylinder (53.0 g). The force exerted on the chiton was calculated as the sum of the buoyant force and the gravitational force. In the case of the wooden cylinder, which floats in water, the chiton was pushed upwards by a force of approximately 0.09 N. For the plastic cylinder, which sinks to the bottom, the chiton sustained a 0.12 N force pulling it downwards. A video of the chiton gripper in action is available on Youtube (https://youtu.be/fL4DzqKwUYw) and as a Supplementary file.

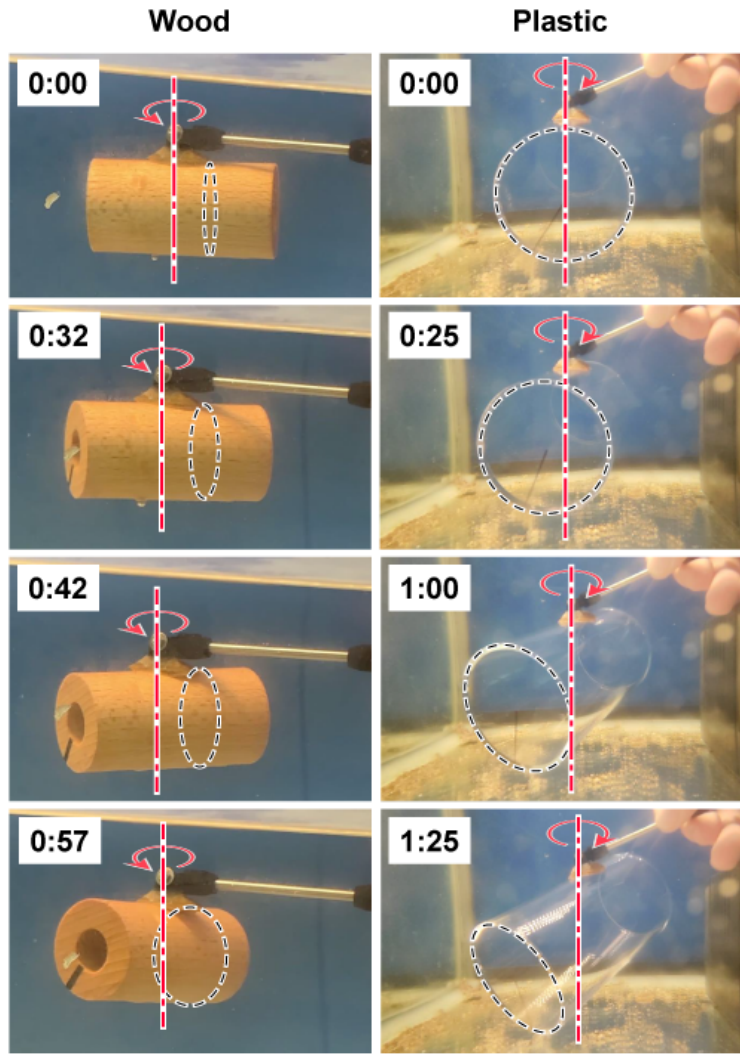

**Figure 8**. Object handling operation using the chiton gripper

**Discussion**

The timing of the grasping left to the reflexive reaction of the pill bug or chiton. Future work shall include the optimization of a method to initiate grasping and release the of the object with more certainty and timing. In the case of the pill bug, the release happened naturally after a given time, however, this was less evident with the chiton. Although the chiton does have the ability to roll into a ball like the pill bug, its preferred mode is sticking to a surface, and it will stick even stronger when stimulated. They have a habit of entering tight spaces to avoid sunlight, and this habit may be exploited. For example, the chiton's ability to move the object may be controlled by optical stimuli such as lasers, and we expect it to be possible to achieve high responsiveness and to drive multiple units independently. Among these methods of stimulation, it is desirable to use the least invasive method possible. In addition, how to reflect and implement changes such as learning and growth of the organism itself in the end-effector in the long run is an important issue in research and development not only from the viewpoint of engineering but also from the viewpoint of investigative science.

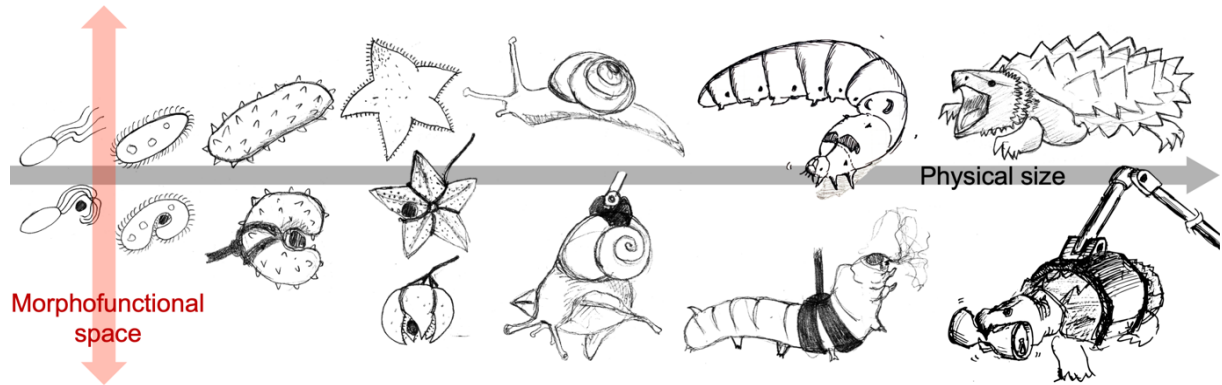

**Figure 9**. Morphofunctional space of biological end effectors.

As an extension of the concept of the open/closed gripper using the pill bug or chiton, we propose a non-exhaustive list of potential methods for grasping manipulation by various organisms, some of which are illustrated in **Figure 9**. A gripper using a living gecko or fly may grasp using the microstructures at the end of their toes. End effectors utilizing the suction cups of octopus, squid, starfish, and frogs are also possible. Furthermore, sea cucumbers can switch between flexible and rigid grasping, and thus can be used as an end effector that switches between soft and rigid modes of grasping. For organisms that adhere to their environment, such as the Japanese spineless mussel or the snail are also useful for sticking to porous objects and objects with rough surfaces. At the microscopic scale, if the distance between the flagella of bacteria can be artificially controlled, it would be possible to switch between swimming and grasping modes to utilize bacteria as micro-handling devices. Ciliates (protozoans) may be used either to generate micro-currents or to collect microscopic objects by endocytosis. The use of swimming and flying animals can also generate water or air flows, respectively, and the animals themselves can carry objects. Spiders or silkworms, which produce silk threads, could be attached to the effector end of a 3D modeling machine to enable the spatial printing of silk. Other potential applications could include a trapping mechanism utilizing the opening and closing actions of carnivorous plants, or leveraging the biting strength of alligator snapping turtles to pierce holes through an object.

The possibilities are endless, but the important point here is that although the body of the organism itself is used as it is, the function is set independently of the intentions of the organism itself. For example, a structure or movement that was intended to be used for locomotion is used for object manipulation, including grasping and moving, or vice versa.

The authors would like to emphasize that no animals were harmed in the process of this research. The pill bug was released back to nature, and the chiton was well active and alive in its aquarium at the time this paper was written, which is exactly six weeks after this experiment, at which point the glued part naturally came off. It will be crucially important to enforce bioethics rules and regulations, especially when dealing with animals that have higher cognition, but since we know little of the cognition of animals in general, the authors recommend caution when handling any type of animal, and to exercise mindfulness in avoiding their suffering as much as possible and to the best of our knowledge. On the other hand, the method to fix or attach each organism should be considered from the engineering viewpoint. Ideally, we should aim for a universal mechanism to reversibly attach various organisms, enabling robots to flexibly make full use of all surrounding elements of their environment, including cooperation with living organisms.

## 5. Conclusion

In this study, from the viewpoint of a mechanism utilizing a living biological organisms' structure and behavior, two end effectors were devised using a living pill bug and chiton. By experimentation with the real prototype models, we confirmed that the grasping and manipulation of very lightweight objects such as cotton was possible for the pill bug, and of relatively heavier objects made of wood or plastic underwater for the chiton. Future challenges will involve improving functionality in a minimally invasive way, including the timing of gripping and release, and the control of the grasping force.

**Acknowledgements**

The authors would like to thank Tsuruoka City Kamo Aquarium for their donation of the live chitons used in this study, and all lab members for their kind support.

**Conflict of Interest**

All authors declare no competing interests.

**Ethical statement**

This study makes use of invertebrates which did not require approval by an ethical committee. Nevertheless, no animals were harmed in the process of this research.